\documentclass[extended-abstract]{ccn}  

\usepackage{amssymb}            
\usepackage{mathtools}          
\usepackage{mathrsfs}           
\mathtoolsset{showonlyrefs}     
\usepackage{graphicx}           
\usepackage{tikz}

\usetikzlibrary{
    positioning,
    arrows.meta,
    shadows,
    fadings,
    shapes.misc
}
\definecolor{sthlmBlue}{RGB}{0, 110, 191}
\definecolor{sthlmRed}{RGB}{196, 0, 100}
\definecolor{sthlmGreen}{RGB}{0, 134, 114}
\definecolor{sthlmGrey}{RGB}{142, 142, 147}
\definecolor{sthlmLightGrey}{RGB}{230, 230, 230}

\usetikzlibrary{positioning} 
\usepackage{subcaption}         
\usepackage[space]{grffile}     
\usepackage{url}                
\usepackage{lipsum}             
\usepackage[space]{grffile}     
\usepackage[english]{babel}

\usepackage{xcolor}             
\usepackage{graphicx}
\usepackage{standalone}

\usepackage{amsmath,amssymb,amsthm,mathtools,amsfonts}
\usepackage{nicefrac}
\usepackage{siunitx}

\usepackage{booktabs}
\usepackage{multirow}
\usepackage{colortbl}
\usepackage{float}
\usepackage{wrapfig}
\usepackage{caption}
\usepackage{enumitem}

\usepackage{pgfplots}
\usepackage{pgfplotstable}
\pgfplotsset{compat=1.18}
\usepgfplotslibrary{fillbetween, groupplots}
\usetikzlibrary{backgrounds, fit, matrix, positioning, shapes.geometric, arrows.meta}

\usepackage{microtype}
\usepackage{url}
\usepackage{hyperref}           
\usepackage{cleveref}           

\definecolor{sthlmBlue}{RGB}{0, 110, 191}
\definecolor{sthlmGrey}{RGB}{142, 142, 147}
\definecolor{sthlmSand}{RGB}{244, 244, 242}
\definecolor{sthlmRedRef}{RGB}{196,0,100} 

\definecolor{sthlmRed}{RGB}{190, 49, 49}
\newcommand{\steady}{\ensuremath{\textcolor{sthlmBlue}{\tau_{steady}}}}
\newcommand{\peak}{\ensuremath{\textcolor{sthlmRed}{\tau_{peak}}}}

\hypersetup{
 linkcolor=sthlmBlue 
,citecolor=sthlmRedRef
,urlcolor=sthlmRedRef
}

\definecolor{findingred}{RGB}{190, 49, 49}

\title{Trace-Mediated Peak Bias: Bridging Temporal Credit Assignment and Cognitive Heuristics in Deep Reinforcement Learning}

\author{
  Viktor Veselý \affmark{1} \And
  Aleksandar Todorov \affmark{1} \And
  Erwan Escudie \affmark{1} \And
  Matthia Sabatelli \affmark{1} 
}
\affiliation{1}{Department of AI, University of Groningen, The Netherlands}

\begin{document}

\maketitle

\begin{abstract}
Temporal credit assignment is central to both biological and artificial intelligence, yet its interaction with non-linear function approximation is poorly understood. We identify a systematic failure mode in deep reinforcement learning (RL) termed Trace-Mediated Peak Bias (TMPB). At intermediate eligibility trace depths, agents irrationally prefer trajectories with high-magnitude reward ``peaks'' over alternatives with higher cumulative returns. This provides a mechanistic account of the Peak-End Rule: a human memory bias where experiences are judged by their most intense moments rather than integrated utility. We show that TMPB emerges because traces amplify distal Temporal Difference errors into ``gradient shocks'' that fixed-step-size Stochastic Gradient Descent cannot normalize, leading to global overestimation. Conversely, adaptive optimizers mitigate this pathology via second-moment normalization. Our results suggest that human-like saliency distortions may emerge naturally from the mathematical constraints of credit assignment in distributed systems, and that adaptive optimization is a theoretical necessity for rational value estimation.
\end{abstract}

\section*{Introduction}
Reinforcement learning (RL) finds its historical and theoretical roots in cognitive psychology \citep{sutton_reinforcement_2020}, descending from Thorndike's Law of Effect to the formalizations of classical conditioning. Central to this lineage is the work of Ivan Pavlov; in his classic experiments, dogs learned to associate a neutral stimulus with a distal reward, a process that inherently requires a biological "trace" to bridge the temporal gap between the bell and the food. In modern RL, this temporal credit assignment, determining which past states are responsible for current rewards, is often solved via eligibility traces \citep{sutton_temporal_1984}. In this framework, an agent seeks to estimate the value function $V^{\pi}(s)$ for a certain state $s$, representing the expected discounted return $G_t = \sum_{k=0}^{\infty} \gamma^k r_{t+k+1}$ for a policy $\pi$, where $\gamma \in [0, 1)$ is the discount factor and $r_t$ is the scalar reward obtained at time-step $t$. The TD($\lambda$) algorithm facilitates this by maintaining an eligibility trace $z_t \in \mathbb{R}^d$, which decays at a rate governed by $\lambda \in [0, 1]$. For a function approximator with parameters $\theta_t$, the update is: $\Delta \theta_t = \alpha \delta_t z_t$ where $\alpha$ is the learning rate and $\delta_t = r_{t+1} + \gamma V_t(s_{t+1}) - V_t(s_t)$ is the temporal difference (TD) error. The trace itself is updated as $z_t = \gamma \lambda z_{t-1} + \nabla_\theta V_t(s_t)$, where $\nabla_\theta V_t(s_t)$ is the gradient of the value estimate with respect to the parameters. Here, $\lambda$ acts as a computational bridge: when $\lambda=0$, the update is purely local (TD(0)); when $\lambda=1$, it mirrors a full Monte Carlo return. While well-defined in tabular settings, traces interact with the shared parameters of non-linear deep neural networks in ways that introduce significant instabilities. We propose that these instabilities manifest as a specific cognitive bias. In psychology, the \textit{Peak-End Rule} describes a heuristic where retrospective evaluations are dominated by the most intense moment (the peak) and the conclusion of an experience, rather than its integrated utility \citep{kahneman2000evaluation}. We identify a direct computational parallel in Deep RL, which we term \textit{Trace-Mediated Peak Bias} (TMPB), suggesting that certain cognitive "irrationalities" may be emergent properties of the mathematical constraints of credit assignment in distributed neural systems.

\section*{Trace Mediated Peak Bias (TMPB)}
To characterize the interaction between eligibility traces and shared neural representations, we conduct a series of policy evaluation experiments. We hypothesize that in non-linear regimes, trace-based credit assignment is susceptible to saliency-driven distortions that mirror human memory heuristics. Specifically, we investigate whether large, distal rewards can "overtake" the value estimates of more frequent but lower-magnitude rewards.

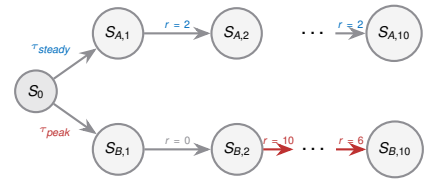
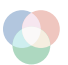
\begin{figure}[htb!] 
    \centering
    \begin{tikzpicture}[
        >=Stealth, 
        node distance=0.8cm and 1.1cm, 
        scale=0.8, transform shape,
        thick,
        state/.style={circle, draw=sthlmGrey, fill=sthlmGrey!10, thick, minimum size=0.6cm, font=\scriptsize},
        rewardStyle/.style={font=\tiny\itshape, inner sep=2pt}, 
        pathLabel/.style={font=\tiny\bfseries, color=sthlmGrey}
    ]
        \node[state, fill=sthlmGrey!20] (s0) {$S_0$};
        \node[state, above right=0.4cm and 0.8cm of s0] (s11) {$S_{A,1}$};
        \node[state, right=of s11] (s12) {$S_{A,2}$}; 
        \node[right=0.5cm of s12] (dots1) {\dots};
        \node[state, right=0.5cm of dots1] (s1T) {$S_{A,10}$};
        \node[state, below right=0.4cm and 0.8cm of s0] (s21) {$S_{B,1}$};
        \node[state, right=of s21] (s22) {$S_{B,2}$};
        \node[right=0.5cm of s22] (dots2) {\dots};
        \node[state, right=0.5cm of dots2] (s2T) {$S_{B,10}$};
        \draw[->, sthlmGrey] (s0) -- node[above left, pathLabel, xshift=2pt] {\steady} (s11);
        \draw[->, sthlmGrey] (s0) -- node[below left, pathLabel, xshift=2pt] {\peak} (s21);
        \draw[->, sthlmGrey] (s11) -- node[above, rewardStyle, color=sthlmBlue] {$r=2$} (s12);
        \draw[->, sthlmGrey] (dots1) -- node[above, rewardStyle, color=sthlmBlue] {$r=2$} (s1T);
        \draw[->, sthlmGrey] (s21) -- node[above, rewardStyle, color=sthlmGrey] {$r=0$} (s22);
        \draw[->, sthlmRed] (s22) -- node[above, rewardStyle, color=sthlmRed, font=\tiny\bfseries] {$r=10$} (dots2);
        \draw[->, sthlmRed] (dots2) -- node[above, rewardStyle, color=sthlmRed, font=\tiny\bfseries] {$r=6$} (s2T);
    \end{tikzpicture}
    \caption{The MDP used for policy evaluation.}
    \label{fig:env_diagram}
\end{figure}

\paragraph{Experimental Setup} We design a stylized episodic Markov Decision Process (MDP) termed the Two-Door Environment (Fig. \ref{fig:env_diagram}). The environment consists of two distinct paths, \steady \: and \peak \:, each of length $T=10$, originating from a common choice point $S_0$. The reward structures contrast uniform density with localized saliency: \steady \: provides a constant reward $r_t = 2$ for all $t \in [1, 10]$, while \peak \: yields a sparse profile with a high-magnitude "peak" $r_3 = 10$, a "concluding" reward $r_{10} = 6$, and zero reward elsewhere. Given $\gamma = 0.95$, the rational values are $V^\star(S_0 | \steady) \approx 16.05$ and $V^\star(S_0 | \peak) \approx 12.81$. We evaluate value estimates $V_{\theta}(s_0)$ using a neural network with one hidden layer of 16 ReLU units. Training is conducted over 1,000 episodes using a uniform behavior policy $\pi_b$ and SGD ($\eta = 0.01$). We utilize the forward-view equivalent of the gradient-based eligibility trace, defining the loss as: $\mathcal{L}(\theta) = \sum_{t=0}^{T-1} \lambda^t ( V_\theta(s_t) - [r_t + \gamma V_\theta(s_{t+1})] )^2$.

\begin{figure}[htb!]
    \centering
    \begin{tikzpicture}
        \begin{groupplot}[
            group style={
                group size=2 by 1,
                horizontal sep=1.2cm, 
            },
            width=0.52\columnwidth, 
            height=4.0cm,
            xlabel=\textbf{{\tiny $\lambda$}},
            ylabel=\textbf{{\tiny $V_0$}},
            ylabel style={yshift=-2pt},
            tick label style={font=\tiny},
            grid=major,
            grid style={sthlmLightGrey, dashed, line width=0.2pt},
            ymin=0, ymax=45, xmin=0, xmax=0.95,
            axis lines=left,
            axis line style={sthlmGrey, thick},
            table/col sep=comma,
            table/x=lambda,
            filter discard if not/.style={
                x filter/.append code={
                    \edef\tmp{\thisrow{optimizer}}
                    \edef\target{#1}
                    \ifx\tmp\target\else\def\pgfmathresult{NaN}\fi
                }
            },
            unbounded coords=discard,
        ]

        \pgfplotsset{
            add ref/.style={
                execute at end plot visualization={
                    \draw[sthlmBlue, thin, dashed, opacity=0.4] (axis cs:0,16.05) -- (axis cs:0.95,16.05);
                    \draw[sthlmRed, thin, dashed, opacity=0.4] (axis cs:0,12.81) -- (axis cs:0.95,12.81);
                }
            }
        }

        \nextgroupplot[title={\scriptsize \textbf{SGD}}, add ref]
        \addplot[name path=A1, sthlmBlue, mark=*, mark size=0.8pt, thick, filter discard if not=SGD]
        table[y=v_steady] {./results/optimizer_results.csv};
        \addplot[name path=B1, sthlmRed, mark=square*, mark size=0.8pt, thick, filter discard if not=SGD]
        table[y=v_peak] {./results/optimizer_results.csv};
        \addplot[sthlmGreen, opacity=0.2]
        fill between[of=A1 and B1, split, 
            every segment no 0/.style={fill=none}, 
            every segment no 1/.style={fill=sthlmGreen}, 
            every segment no 2/.style={fill=none}];

        \nextgroupplot[title={\scriptsize \textbf{RMSprop}}, add ref, ylabel={}]
        \addplot[name path=A2, sthlmBlue, mark=*, mark size=0.8pt, thick, filter discard if not=RMSprop]
        table[y=v_steady] {./results/optimizer_results.csv};
        \addplot[name path=B2, sthlmRed, mark=square*, mark size=0.8pt, thick, filter discard if not=RMSprop]
        table[y=v_peak] {./results/optimizer_results.csv};
        \addplot[sthlmGreen, opacity=0.2]
        fill between[of=A2 and B2, split,
            every segment no 1/.style={fill=sthlmGreen}];

        \end{groupplot}

        \node[anchor=north, yshift=-0.6cm] at (current bounding box.south) {
            \begin{tikzpicture}
            \begin{axis}[
                hide axis, xmin=0,xmax=1,ymin=0,ymax=1,
                legend style={draw=none, font=\tiny, legend columns=3, column sep=8pt}
            ]
            \addlegendimage{sthlmBlue, thick, mark=*, mark size=1pt}
            \addlegendentry{Steady}
            \addlegendimage{sthlmRed, thick, mark=square*, mark size=1pt}
            \addlegendentry{Peak}
            \addlegendimage{sthlmGreen, area legend, fill=sthlmGreen, opacity=0.3}
            \addlegendentry{Irrationality Zone ($V_B > V_A$)}
            \end{axis}
            \end{tikzpicture}
        };
    \end{tikzpicture}
    \caption{Policy evaluation results for SGD (left) and RMSprop (right). The Irrationality Zone \textcolor{sthlmGreen}{(green)} vanishes under adaptive optimization.}
    \label{fig:comparison_results}
\end{figure}
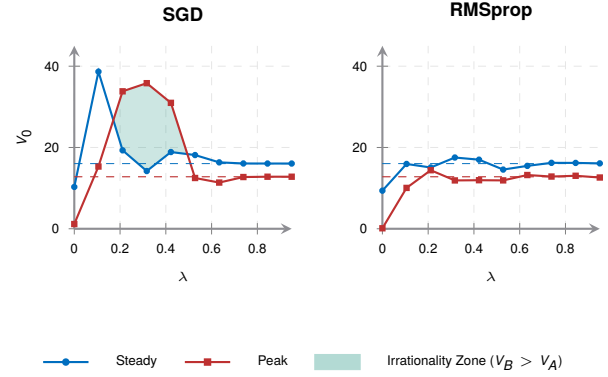

\paragraph{Results} As shown in the first plot of Fig. \ref{fig:comparison_results}, the interaction between eligibility traces and neural function approximation induces a systematic overestimation of $V(S_0 | \peak)$. The sub-optimal preference for the Peak trajectory is most acute at intermediate trace depths ($0.15 < \lambda < 0.50$), where the value estimation for the inferior path erroneously ``leapfrogs'' that of the Steady path. We formally define this phenomenon as Trace-Mediated Peak Bias (TMPB): a computational artifact inspired by the \textit{Peak-End Rule} widely observed in cognitive psychology \citep{dutta_retention_1972, miron-shatz_evaluating_2009}. Just as humans tend to favor experiences with high-intensity peak rewards over those with greater aggregate utility, reinforcement learning agents equipped with eligibility traces and non-linear function approximation exhibit a strikingly similar ``irrational'' preference. 

\paragraph{Mechanistic Account} TMPB arises mechanistically because high-magnitude rewards generate gradient shocks ($\Delta \theta_t = \alpha \delta_t z_t$) that disproportionately reconfigure shared weights. In fixed-step-size SGD, the parameter shift is directly proportional to reward saliency; thus, a single peak update can exceed the integrated sum of steady updates. This suggests a tension between integrative and saliency-driven credit assignment: while an ideal observer integrates reward density, saliency shocks can effectively hijack the neural manifold. Adaptive optimizers (e.g., RMSprop \citep{tieleman2012lecture}) mitigate this (see right plot of Fig. \ref{fig:comparison_results}) by scaling updates by the inverse of the second moment of gradients, dampening shocks as high-variance outliers. This implies that biological Peak-End effects may reflect a reliance on gradient-based updates that lack the variance-normalization seen in adaptive artificial systems.

\section{Discussion \& Conclusion}
We have provided a mechanistic account of how cognitive-like biases can emerge from the fundamental mathematics of temporal credit assignment. Our main contributions are threefold: (1) we identify and formalize Trace-Mediated Peak Bias (TMPB) as a systematic failure mode in deep RL; (2) we demonstrate that this bias provides a computational parallel to the psychological Peak-End Rule; and (3) we show that adaptive optimization is a necessary structural requirement for rational value estimation in non-linear systems. The historical development of reinforcement learning is deeply intertwined with cognitive psychology. Our findings suggest that this relationship remains a powerful two-way street. While psychology originally inspired the RL framework, the mathematical pathologies of modern deep RL now provide a rigorous language for explaining human "irrationalities." The Peak-End Rule has long been viewed as an isolated psychological heuristic; however, we demonstrate that it is an emergent property of any system utilizing distributed representations and trace-like credit assignment without variance-based normalization. This interaction offers a significant lesson for both artificial and biological intelligence. For AI, it highlights that "human-like" neural architectures naturally inherit human-like vulnerabilities, making the choice of optimizer a theoretical concern rather than a mere hyperparameter. For cognitive science, it suggests that human saliency distortions may reflect a biological reliance on simple gradient-based updates that lack the sophisticated normalization found in adaptive methods. 

\printbibliography

\end{document}